\pdfoutput=1

\documentclass[11pt]{article}

\usepackage[final]{acl}

\usepackage{times}
\usepackage{latexsym}

\usepackage[T1]{fontenc}
\usepackage{amssymb} 
\usepackage[utf8]{inputenc}

\usepackage{microtype}

\usepackage{inconsolata}

\usepackage{graphicx}
\usepackage{tabularx}
\usepackage{booktabs}
\usepackage{multirow}
\usepackage{xspace}
\newcommand{\mymethod}{MemInsight\xspace}

\usepackage{amsmath}

 \author{\small Rana Salama, Jason Cai, Michelle Yuan, Anna Currey, Monica Sunkara, Yi Zhang, Yassine Benajiba
\\ 
\small AWS AI\\
 \small\texttt{\{ranasal, cjinglun, miyuan, ancurrey, sunkaral, yizhngn, benajiy
 \}}@amazon.com
 }

\title{\mymethod: Autonomous Memory Augmentation for LLM Agents}

\begin{document}

\maketitle

\begin{abstract}

Large language model (LLM) agents have evolved to intelligently process information, make decisions, and interact with users or tools. A key capability is the integration of long-term memory capabilities, enabling these agents to draw upon historical interactions and knowledge. However, the growing memory size and need for semantic structuring pose significant challenges. In this work, we propose an autonomous memory augmentation approach, \mymethod, to enhance semantic data representation and retrieval mechanisms. By leveraging autonomous augmentation to historical interactions, LLM agents are shown to deliver more accurate and contextualized responses. We empirically validate the efficacy of our proposed approach in three task scenarios; conversational recommendation, question answering and event summarization. On the LLM-REDIAL dataset, \mymethod boosts persuasiveness of recommendations by up to 14\%. Moreover, it outperforms a RAG baseline by 34\% in recall for LoCoMo retrieval. Our empirical results show the potential of \mymethod to enhance the contextual performance of LLM agents across multiple tasks

\end{abstract}

\section{Introduction}

LLM agents have emerged as an advanced framework to extend the capabilities of LLMs to improve reasoning~\cite{react,rethink}, adaptability~\cite{adapt}, and self-evolution~\cite{expel,self_ev,self_critic}. A key component of these agents is their memory module, which retains past interactions to allow more coherent, consistent, and personalized responses across various tasks. The memory of the LLM agent is designed to emulate human cognitive processes by simulating how knowledge is accumulated and historical experiences are leveraged to facilitate complex reasoning and the retrieval of relevant information to inform actions~\cite{survey}. However, the advantages of an LLM agent's memory also introduce notable challenges~\cite{survey2}. 
As interactions accumulate over time, retrieving relevant information becomes increasingly difficult, especially in long-term or complex tasks. Raw historical data grows rapidly and, without effective memory management, can become noisy and imprecise, hindering retrieval and degrading agent performance. Moreover, unstructured memory limits the agent’s ability to integrate knowledge across tasks and contexts. Therefore, structured knowledge representation is essential for efficient retrieval, enhancing contextual understanding, and supporting scalable long-term memory in LLM agents. Improved memory management enables better retrieval and contextual awareness, making this a critical and evolving area of research.

Hence, in this paper we introduce an autonomous memory augmentation approach, \mymethod, which empowers LLM agents to identify critical information within the data and proactively propose effective attributes for memory enhancements. This is analogous to the human processes of attentional control and cognitive updating, which involve selectively prioritizing relevant information, filtering out distractions, and continuously refreshing the mental workspace with new and pertinent data~\cite{himem,myagent}.

\mymethod autonomously generates augmentations that encode both relevant semantic and contextual information for memory. These augmentations facilitate the identification of memory components pertinent to various tasks. Accordingly, \mymethod can improve memory retrieval by leveraging relevant attributes of memory, thereby supporting autonomous LLM agent  adaptability and self-evolution.
\\
\\
Our contributions can be summarized as follows:
\begin{itemize}
\item We propose a structured autonomous approach that adapts LLM agents' memory representations while preserving context across extended conversations for various tasks.
\item We design and apply memory retrieval methods that leverage the generated memory augmentations to filter out irrelevant memory while retaining key historical insights.
\item Our promising empirical findings demonstrate the effectiveness of \mymethod on several tasks: conversational recommendation, question answering, and event summarization.
\end{itemize}

\section{Related Work}
Well-organized and semantically rich memory structures enable efficient storage and retrieval of information, allowing LLM agents to maintain contextual coherence and provide relevant responses. Developing an effective memory module in LLM agents typically involves two critical components: structural memory generation and memory retrieval methods~\cite{survey,mix_agents}.

\paragraph{LLM Agents Memory}
Recent research in LLM agents memory focuses on storing and retrieving prior interactions to improve adaptability and generalization~\cite{memgpt,expel,survey,gim}. Common approaches structure memory as summaries, temporal events, or reasoning chains to reduce redundancy and highlight key information~\cite{locomo,episodic,tim}. Some methods enrich raw dialogues with semantic annotations, such as event sequences~\cite{memorybank,locomo} or reusable workflows~\cite{workflow}.
Recent models like A-Mem\cite{amem} that uses manually defined task-specific notes to structure an agent's memory, while Mem0\cite{mem0} offers a scalable, real-time memory pipeline for production use. However, most existing methods rely on unstructured memory or manually defined schemas. In contrast, \mymethod autonomously discovers semantically meaningful attributes, enabling structured memory representation without human-crafted definitions.

\paragraph{LLM Agents Memory Retrieval}

Recent work has explored memory retrieval techniques to improve efficiency when handling large-scale historical context in LLM agents~\cite{hu2023chatdb,zhao2024expel,tack2024online,ge2025tremu}. Common approaches involves generative retrieval models, which encode memory entries as dense vectors and retrieve the top-$k$ most relevant documents using similarity search~\cite{memorybank,bridge}. Similarity metrics such as cosine similarity~\cite{memgpt} are widely used, often in combination with dual-tower dense retrievers, where memory entries are embedded independently and indexed via tools like FAISS~\cite{faiss} for efficient retrieval~\cite{memorybank}. Additionally, techniques such as Locality-Sensitive Hashing (LSH) are utilized to retrieve tuples containing related entries in memory~\cite{chatdb}.

\section{Autonomous Memory Augmentation}

\begin{figure*}
    \centering
\includegraphics[width=0.85\linewidth]{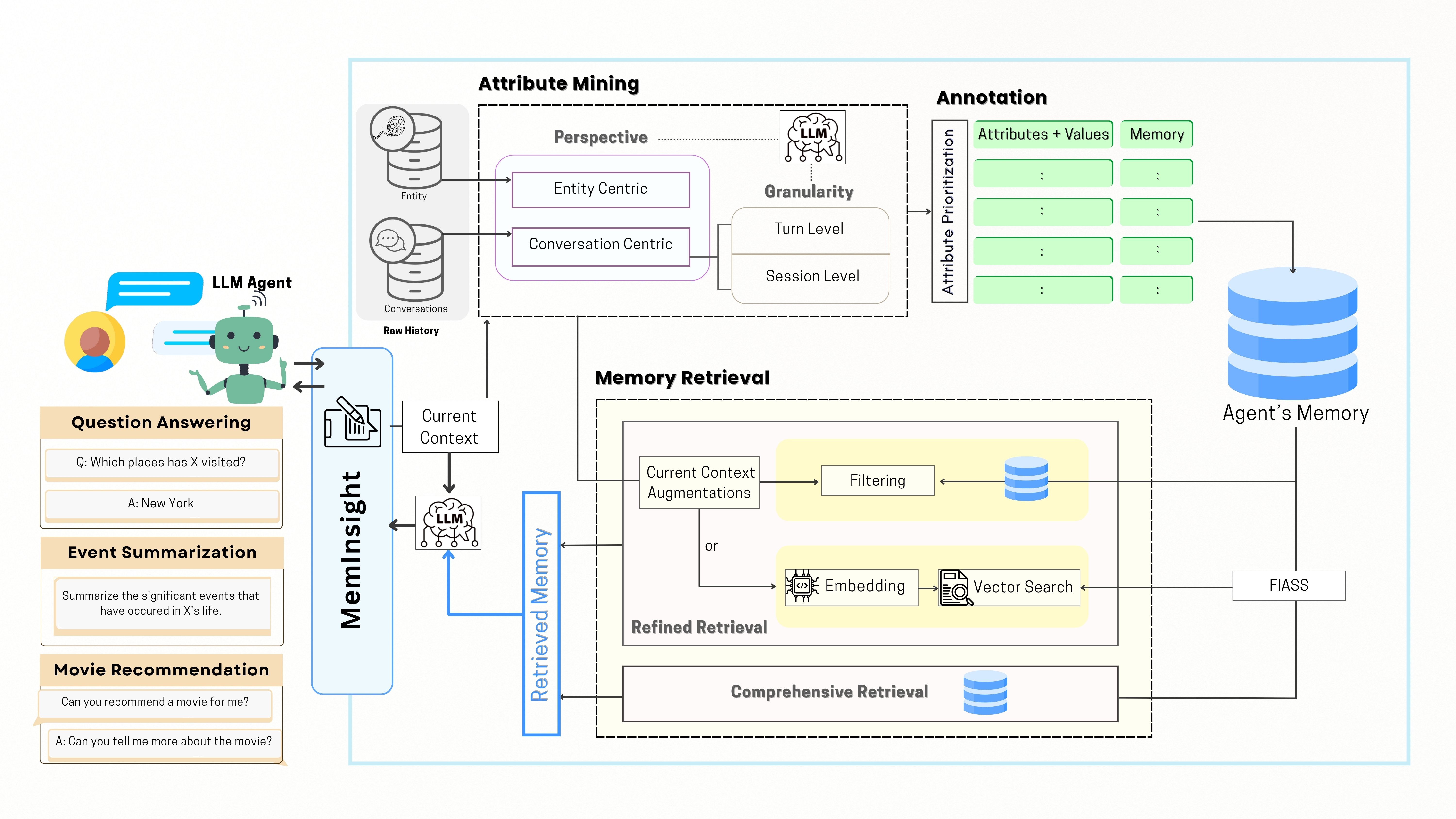}\small \caption{ \small \mymethod framework comprising three core modules: Attribute Mining (including perspective and granularity), Annotation (with attribute prioritization), and Memory Retrieval (including refined and comprehensive retrieval). These components are triggered by various downstream tasks such as Question Answering, Event Summarization, and Conversational Recommendation.\vspace{-2mm}}
    \label{model}
\end{figure*}
Our proposed model, \mymethod, encapsulates the agent's memory $M$, offering a unified framework for augmenting and retrieving user–agent interactions represented as memory instances $m$.
As new interactions occur, they are autonomously augmented and incorporated into memory, forming an enriched set $M = \{m_{1{<augmented>}}, \dots, m_{n{<augmented>}}\}$. As shown in Figure~\ref{model}, \mymethod comprises three core modules: attribute mining, annotation, and memory retrieval.

\subsection {Attribute Mining and Annotation}

Attribute mining  extracts structured and semantically meaningful attributes from input dialogues for memory augmentation. The process follows a principled framework guided by three key dimensions: 
\\(1) \textit{Perspective}, from which attributes are derived (e.g., entity- or conversation-centric annotations)
\\(2) \textit{Granularity}, indicating the level of annotation detail (e.g., turn-level or session-level)
\\(3) \textit{Annotation}, which ensures that extracted attributes are appropriately aligned with the corresponding memory instance. A backbone LLM is leveraged to autonomously identify and generate relevant attributes. 

\subsubsection {Attribute Perspective} 

An attribute perspective entails two main orientations: entity-centric and conversation-centric. The entity-centric focuses on annotating specific items referenced in memory, such as books or movies, using attributes that capture their key properties (e.g., director, author, release year). In contrast, the conversation-centric perspective captures attributes that reflect the overall user interaction with respect to users' intent, preferences, sentiment, emotions, motivations, and choices, thereby improving response generation and memory retrieval. An illustrative example is provided in Figure \ref{fig:aug_ex}.

\begin{figure} []
    \centering
    \includegraphics[width=1\linewidth]{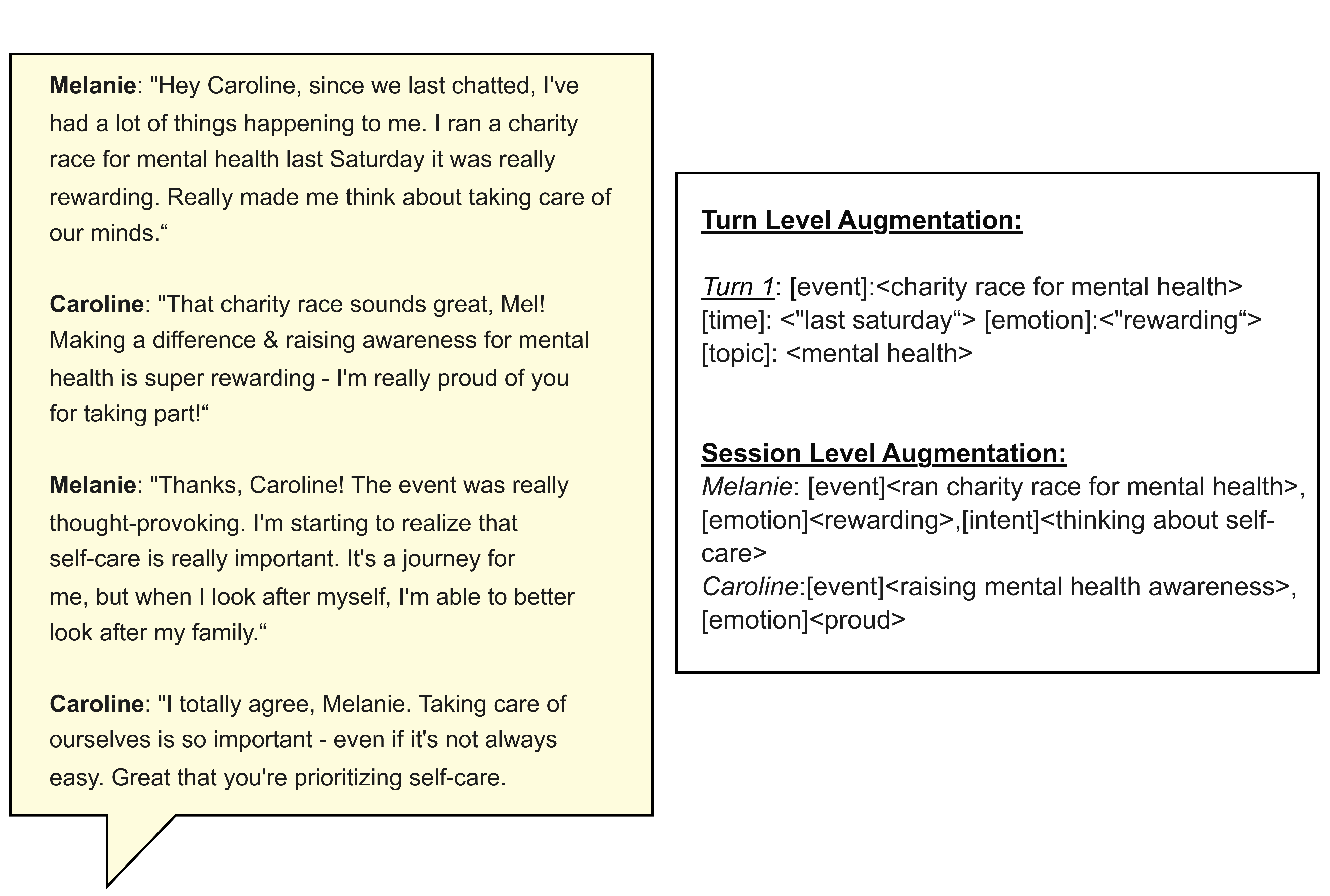}
    \small \caption{ \small An example for Turn level and Session level annotations for a sample dialogue conversation from the LoCoMo Dataset. \vspace{-2mm} }
    \label{fig:tl_sl}
\end{figure}

\subsubsection{Attribute Granularity}

Conversation-centric augmentations introduce the notion of attribute granularity, which defines the level of details captured in the augmentation process. The augmentation attributes can be analyzed at varying levels of abstraction, either at the level of individual turns within a user conversation (turn-level), or across the entire dialogue session (session-level), each offering distinct insights into the conversational context. Turn-level focuses on the specific content of individual turns to generate more nuanced and contextual attributes, while session-level augmentation captures broader patterns and user intent across the interaction. Figure~\ref{fig:tl_sl} illustrates this distinction, showing how both levels offer complementary perspectives on a sample dialogue.

\subsubsection{Annotation and Attribute Prioritization}
Subsequently, the generated attributes and their corresponding values are used to annotate the agent's memory. Annotation is done by aggregating attributes and values in the relevant memory. 
\\Given an interaction $i$, the module applies an LLM-based extraction function $\mathcal{F}_{\text{LLM}}$ to produce a set of attribute–value pairs:
\[
A = \mathcal{F}_{\text{LLM}}(D) = \{ (a_j, v_j) \}_{j=1}^k
\]
where: \( a_j \in \mathcal{A} \) represents the attribute (e.g., emotion, entity, intent) and \( v_j \in \mathcal{V} \) the value of this attribute
These attributes are then used to annotate the corresponding memory instance $m_i$, resulting in an augmented memory $M_a$:
$M_a = \{(A_1, \tilde{m}_1), (A_2, \tilde{m}_2), \dots, ( A_i, \tilde{m}_i)\},$
Attributes are typically aggregated using the Attribute Prioritization method, which can be classified into Basic and Priority. In Basic Augmentation, attributes are aggregated without a predefined order, resulting in an arbitrary sequence. In contrast, Priority Augmentation sorts attribute-value pairs according to their relevance to the memory being augmented. This prioritization follows a structured order in which attribute $(A_1, \tilde{m_1})$ holds the highest significance, ensuring that more relevant attributes are processed first.

\subsection {Memory Retrieval}


\mymethod augmentations are employed to both enrich memory representations and support the retrieval of contextually relevant memory. These augmentations are utilized in one of two ways. 
\\(1) Comprehensive retrieval, retrieves all related memory instances along with their associated augmentations to support context-aware inference. 
\\(2) Refined retrieval, where the current context is augmented to extract task-specific attributes, which then guide the retrieval process through one of the following methods:
\\a- \textit{Attribute-based Retrieval}: which uses the current attributes as filters to select memory instances with matching or related augmentations only. 
Given a query session \( Q \) with attributes \( A_Q \), retrieve relevant memories:
\begin{equation*}
\resizebox{0.45\textwidth}{!}{$\mathcal{R}_{\text{attr}}(A_Q, \mathbb{M}) = \text{Top-}k \left\{ (A_k,M_k) \mid \text{match}(A_Q, A_k) \right\}$}
\end{equation*}
b- \textit{Embedding-based Retrieval} where memory augmentations are embedded as dense vectors. A query embedding is derived from the current context’s augmentations and used to retrieve the top-$k$ most similar memory entries via similarity search. 
Let \( \phi: A_k \rightarrow \mathbb{V}^d \) be the embedding function over attributes. Then:
\begin{equation*}
\resizebox{0.35\textwidth}{!}{$\footnotesize{\textit{$sim$}(A_Q, A_k)} = \frac{\phi(A_Q) \cdot \phi(A_k)}{\|\phi(A_Q)\| \cdot \|\phi(A_k)\|}$}
\end{equation*}
\begin{equation*}
\resizebox{0.45\textwidth}{!}{$\mathcal{R}_{\text{embed}}(A_Q, \mathbb{M}) = \text{Top-}k \left\{ (A_k,M_k) \mid \text{sim}(A_Q, A_k) \right\}$}
\end{equation*}
Finally, the retrieved memories are then integrated into the current context to inform the ongoing interaction.
Further implementation details of embedding-based retrieval are provided in Appendix~\ref{emb_ret}.

\section{Evaluation}
\subsection{Datasets}
We evaluate \mymethod on two benchmarks: LLM-REDIAL~\cite{llm_redial} and LoCoMo~\cite{locomo}. LLM-REDIAL is a dataset for conversational movie recommendation, comprising ~10K dialogues and 11K movie mentions. LoCoMo is a dataset for evaluating Question Answering and Event Summarization, with 30 multi-session dialogues between two speakers. It features five question types: Single-hop, Multi-hop, Temporal reasoning, Open-domain, and Adversarial, each annotated with the relevant dialogue turn required for answering. LoCoMo also provides event labels for each speaker in a session, which serve as ground truth for evaluating event summarization.

\subsection{Experimental Setup}
To evaluate our model, we begin by augmenting the datasets using zero-shot prompting to extract relevant attributes and their corresponding values. For attribute generation across tasks, we employ Claude Sonnet\footnote{\scriptsize claude-3-sonnet-20240229-v1}, LLaMA 3\footnote{\scriptsize llama3-70b-instruct-v1}, and Mistral\footnote{\scriptsize mistral-7b-instruct-v0}. For the Event Summarization task, we additionally utilize Claude 3 Haiku\footnote{\scriptsize claude-3-haiku-20240307-v1}. In embedding-based retrieval, we use the Titan Text Embedding model\footnote{\scriptsize titan-embed-text-v2:0} to generate embeddings of augmented memory, which are indexed and searched using FAISS~\cite{faiss}.
To ensure consistency across all experiments, we use the same base model for the primary tasks: recommendation, answer generation, and summarization, while varying the models used for memory augmentation. Claude Sonnet serves as the backbone LLM in all baseline evaluations.

\subsection{Evaluation Metrics}
We evaluate \mymethod using a combination of standard and LLM-based metrics. For Question Answering, we report F1-score for answer prediction and recall for accuracy; for Conversational Recommendation, we use Recall@K, NDCG@K, along with LLM-based metrics for genre matching. 
\\We further incorporate subjective metrics, including \textit{Persuasiveness} ~\cite{llm_redial}, which measures how persuasive a recommendation aligns with the ground truth. Additionally, we introduce a \textit{Relatedness} metric where we prompt an LLM to measure how comparable are recommendation attributes to the ground truth, categorizing them as not comparable, comparable, or highly comparable.
For Event Summarization, we adopt G-Eval~\cite{geval}, an LLM-based metric that evaluates the relevance, consistency, and coherence of generated summaries against reference labels. 
Together, these metrics provide a comprehensive framework for evaluating both retrieval effectiveness and response quality.

\section{Experiments}
\begin{table*}[]
\centering
\scriptsize
\begin{tabular}{lcccccc}
\hline
\multicolumn{1}{c}{ \textbf{Model}}         &  \textbf{Single-hop}         &  \textbf{Multi-hop}          &  \textbf{Temporal} & \textbf{Open-domain} &  \textbf{Adversarial}      &  \textbf{Overall}       \\   \hline
\multicolumn{1}{l}{Baseline (Claude-3-Sonnet)} 
& 15.0  & 10.0  & 3.3 & 26.0   & 45.3  & 26.1 \\

\multicolumn{1}{l}{LoCoMo (Mistral v1)} 
& 10.2  & \textbf{12.8} &\textbf{ 16.1} & 19.5   & 17.0  & 13.9 \\
      
\multicolumn{1}{l}{ReadAgent (GPT-4o)} 
& 9.1  & 12.6  & 5.3 & 9.6  & 9.81  & 8.5 \\

\multicolumn{1}{l}{MemoryBank (GPT-4o)} 
& 5.0 & 9.6 & 5.5 & 6.6  & 7.3 & 6.2\\
\hline
\multicolumn{7}{l}{\textbf{Attribute-based Retrieval}} \\ 
\mymethod (Claude-3-Sonnet)  & \textbf{18.0 } & 10.3 & 7.5 & \textbf{27.0 } & \textbf{58.3} & \textbf{29.1} \\
\hline
\hline
\multicolumn{7}{l}{\textbf{Embedding-Based Retrieval}} \\ 
\multicolumn{1}{l}{ RAG Baseline (DPR)} & 11.9   & 9.0   & 6.3  & 12.0   & \textbf{89.9} & 28.7 
\\\hline
 \mymethod (Llama v3$_{Priority}$) & 14.3  & 13.4  & 6.0   & 15.8  & 82.7 & 29.7   \\
\mymethod (Mistral v1$_{Priority}$) &  \textbf{16.1} & 14.1 & 6.1 &  16.7 & 81.2  & 30.0  \\\hline
\mymethod (Claude-3-Sonnet$_{Basic}$) & 14.7  & 13.8  & 5.8  & 15.6  & 82.1 & 29.6   
\\
\mymethod (Claude-3-Sonnet$_{Priority}$)  & 15.8   & \textbf{15.8}  & 6.7 & \textbf{19.7}   & 75.3  & \textbf{30.1}
\\\hline
\end{tabular}
\small \caption{ \label{qa_f1} \small Results for F1 Score (\%) for answer generation accuracy for attribute-based and embedding-based memory retrieval methods. Baseline is Claude-3-Sonnet model to generate answers using all memory without augmentation, for Attribute-based retrieval. In addition to the Dense Passage Retrieval(DPR) for Embedding-based retrieval. Evaluation is done with $k=5$. Best results per question category over all methods are in bold.}
\end{table*}

\begin{table*}[]
\centering
\scriptsize
\begin{tabular}{lcccccc}
\hline
\multicolumn{1}{c}{ \textbf{Model}}         &  \textbf{Single-hop }        &  \textbf{Multi-hop}          &  \textbf{Temporal} & \textbf{Open-domain} &  \textbf{Adversarial}      &  \textbf{Overall}        \\ \hline
RAG Baseline (DPR)   & 15.7   & 31.4    & 15.4  & 15.4   & 34.9  & 26.5
\\

\mymethod (Llama v3$_{Priority}$)                 & 31.3                & 63.6               & 23.8      & 53.4      & 28.7                 & 44.9                  \\
\mymethod (Mistral v1$_{Priority}$)  & 31.4                & 63.9               & 26.9                                                                  & 58.1                                                                     & 36.7                 & 48.9           \\
\mymethod (Claude-3-Sonnet$_{Basic}$) & 33.2   & 67.1  & 29.5   & 56.2 & 35.7 & 48.8           \\
\mymethod (Claude-3-Sonnet$_{Priority}$) & \textbf{39.7}       & \textbf{75.1}  & \textbf{32.6} & \textbf{70.9} & \textbf{49.7}        & \textbf{60.5}        
      \\ \hline
\end{tabular}
\small \caption{ \label{qa_recall} \small Results for the RECALL@k=5 accuracy for Embedding-based  retrieval for answer generation using LoCoMo dataset. Dense Passage Retrieval(DPR) RAG model is the baseline. Best results are in bold. \vspace{-1mm}}
\end{table*}

\subsection{Questioning Answering}

Question Answering experiments are conducted to evaluate the effectiveness of \mymethod in answer generation. We evaluate the overall accuracy to measure the system’s ability to retrieve and integrate relevant information using memory augmentations. The base model, which incorporates all historical dialogues without any augmentation, serves as a baseline. Additionally, we report results on the LoCoMo benchmark using the same backbone model (Mistral v1) to ensure a fair evaluation. We also compare with stronger GPT-based baselines, including MemoryBank\cite{memorybank} and ReadAGent~\cite{readagent}, which utilizes external memory modules to support long-term reasoning. We also consider Dense Passage Retrieval (DPR)~\cite{dpr} as a representative baseline of RAG due to its scalability and retrieval efficiency.

\paragraph{Memory Augmentation}
In this task, memory is constructed from historical conversational dialogues, which requires the generation of conversation-centric attributes for augmentation. Given that the ground-truth labels consist of dialogue turns relevant to the question, the dialogues are annotated at the turn level. An LLM backbone is prompted to generate augmentation attributes for both conversation-centric and turn-level annotations.

\paragraph{Memory Retrieval}
To answer a given question, \mymethod first augments it to extract relevant attributes, which guide memory retrieval. In attribute-based retrieval, dialogue turns with matching augmentation attributes are retrieved. In embedding-based retrieval, the question and its attributes are embedded to perform a vector similarity search over indexed memory. The top-$k$ most similar dialogue turns are then integrated into the current context to generate an answer.

\paragraph{Experimental Results}

As shown in Table~\ref{qa_f1}, \mymethod achieves significantly higher overall accuracy on the question answering task compared to all baselines, using both attribute-based and embedding-based memory retrieval. In the attribute-based setting, \mymethod with Claude-3-Sonnet demonstrates notable gains in single-hop, temporal, and adversarial questions, which require more complex contextual reasoning. These results highlight the effectiveness of memory augmentation in enriching context and enhancing answer quality. \mymethod further outperforms all other benchmark models across most question types, with the exception of multi-hop and temporal questions in LoCoMo, where evaluation is based on a partial-match F1 metric~\cite{locomo}.

For embedding-based retrieval, we evaluate \mymethod using both basic and priority augmentation, alongside the DPR baseline. \mymethod consistently outperforms all baselines, except in temporal and adversarial questions, where DPR achieves slightly higher accuracy. Nevertheless, \mymethod maintains the highest overall accuracy. Priority augmentation also consistently outperforms basic augmentation across nearly all question types, validating its effectiveness in improving contextual relevance. Notably, \mymethod demonstrates substantial gains on multi-hop questions, which require reasoning over multiple pieces of supporting evidence, highlighting its ability to integrate dispersed information from historical dialogue. As shown in Table~\ref{qa_recall}, recall metrics further support this trend, with priority augmentation yielding a 35\% overall improvement and consistent gains across all categories.

\begin{table}[]
\footnotesize
\centering
\begin{tabular}{llll}
\hline
\textbf{Statistic}                       &  & \multicolumn{2}{l}{\textbf{Count}}  \\ \hline
Total Movies                     & & \multicolumn{2}{l}{9687}   \\
Avg. Attributes                  & & \multicolumn{2}{l}{7.39}   \\
Failed Attributes                & & \multicolumn{2}{l}{0.10\%} \\
\hline
\multirow{5}{*}{Top-5 Attributes} & Genre            & 9662    \\
                                  & Release year     & 5998    \\
                                  & Director         & 5917    \\
                                  & Setting          & 4302    \\
                                  & Characters       & 3603    \\ \hline
\end{tabular}
\small \caption{\label{movie_att} 
\small Statistics of attributes generated for the LLM-REDIAL Movie dataset, which include total number of movies, average number of attributes per item, number of failed attributes, and the counts for the most frequent five attributes.
\vspace{-2mm}}
\end{table}

\begin{table*}[]
\centering
\scriptsize
\begin{tabular}{lcccccccccc}
\toprule
\multicolumn{1}{c}{\textbf{Model}} & \begin{tabular}[c]{@{}c@{}}\textbf{Avg. Items} \\ \textbf{Retrieved}\end{tabular} & \multicolumn{3}{c}{\textbf{Direct Match (↑)}}                         & \multicolumn{3}{c}{\textbf{Genre Match (↑)}}                                                    & \multicolumn{3}{c}{\textbf{NDCG}(↑)}                                \\ 
\midrule
   & \multicolumn{1}{l}{}                                                                  & R@1  & R@5             & R@10             & R@1            & \multicolumn{1}{c}{R@5} & R@10                     & N@1  & N@5             & N@10            \\ 
\midrule
Baseline (Claude-3-Sonnet)  & 144   & 0.000  & 0.010  & 0.015 & 0.320  & 0.57  & 0.660  & 0.005 & 0.007 & 0.008                   \\
LLM-REDIAL Model & 144 & - & 0.000  & 0.005  & - & -  & -  & - & 0.000 & 0.001
\\
\hline
\multicolumn{11}{l}{\textbf{Attribute-Based Retrieval}} 
\\
\mymethod (Claude-3-Sonnet) & 15  & 0.005  & 0.015 & 0.015 & 0.270 & 0.540 & 0.640 & 0.005 & 0.007 & 0.007
\\
\hline
\multicolumn{11}{l}{\textbf{Embedding-Based Retrieval}} \\
\mymethod (Llama v3) & 10 & 0.000  & 0.005  & \textbf{0.028} & 0.380  & 0.580 & 0.670 & 0.000 & 0.002          & 0.001
\\
\mymethod (Mistral v1) & 10 & 0.005 & 0.010 & 0.010  & 0.380  & 0.550 & 0.630  & 0.005 & 0.007 & 0.007 
\\
\mymethod (Claude-3-Haiku) & 10  & 0.005 & 0.010          & 0.010  & 0.360 & \textbf{0.610} & 0.650 & 0.005 & 0.007 & 0.007 
\\
\mymethod (Claude-3-Sonnet) & 10 & 0.005 & 0.015 & 0.015 & \textbf{0.400} & 0.600 & 0.64 & 0.005 & 0.010 & 0.010 
\\
\hline
\multicolumn{11}{l}{\textbf{Comprehensive}} \\
\mymethod (Claude-3-Sonnet) & 144  & \textbf{0.010} & \textbf{0.020} & 0.025  & 0.300   & 0.590 & \textbf{0.690}  & \textbf{0.010} & \textbf{0.015}  & \textbf{0.017}
\\
\bottomrule
\end{tabular}
\small \caption{ \label{cr_res} \small Results for Movie Conversational Recommendation using (1) Attribute-based retrieval with Claude-3-Sonnet model (2) Embedding-based retrieval across models (Llama v3, Mistral v1, Claude-3-Haiku, and Claude-3-Sonnet) (3) Comprehensive setting using Claude-3-Sonnet that includes \textbf{ALL} augmentations. Evaluation metrics include RECALL, NDCG, and an LLM-based genre matching metric, with $n=200$ and $k=10$. Baseline is Claude-3-Sonnet without augmentation. Best results are in bold.
\vspace{-2mm}
}
\end{table*}



\subsection{Conversational Recommendation}
We simulate conversational recommendation by preparing dialogues for evaluation under the same conditions proposed by~\citet{llm_redial}. This process involves masking the dialogue and randomly selecting $n=200$ conversations for evaluation to ensure a fair comparison. Each conversational dialogue used is processed by masking the ground truth labels, followed by a turn cut-off, where all dialogue turns following the first masked turn are removed and retained as evaluation labels. Subsequently, the dialogues are augmented using a conversation-centric approach to identify relevant user interest attributes for retrieval.
Finally, we prompt the LLM model to generate a movie recommendation that best aligns with the masked token, guided by the augmented movies retrieved based on the user’s historical interactions.

The baseline for this evaluation is the results presented in the LLM-REDIAL paper~\cite{llm_redial} which employs zero-shot prompting for recommendation using the ChatGPT model\footnote{https://openai.com/blog/chatgpt}. In addition to the baseline model that uses memory without augmentation. 

Evaluation includes direct matches between recommended and ground truth movie titles using RECALL@[1,5,10] and NDCG@[1,5,10]. Furthermore, to address inconsistencies in movie titles generated by LLMs, we incorporate an LLM-based evaluation that assesses recommendations based on genre similarity. Specifically, a recommended movie is considered a  valid match if it shares the same genre as the corresponding ground truth label.

\paragraph{Memory Augmentation}
We initially augment the dataset with relevant attributes, primarily employing entity-centric augmentations for memory annotation, as the memory consists of movies.
In this context, we conduct a detailed evaluation of the generated attributes to provide an initial assessment of the effectiveness and relevance of \mymethod augmentations. To evaluate the quality of the generated attributes, Table \ref{movie_att} presents statistical data on the generated attributes, including the five most frequently occurring attributes across the entire dataset.
As shown in the table, the generated attributes are generally relevant, with "genre" being the most significant attribute based on its cumulative frequency across all movies (also shown in Figure \ref{fig:all_att}). However, the relevance of attributes vary, emphasizing the need for prioritization in augmentation. Additionally, the table reveals that augmentation was unsuccessful for 0.1\% of the movies, primarily due to the LLM's inability to recognize certain movie titles or because the presence of some words in the movie titles conflicted with the LLM's policy.

\paragraph{Memory Retrieval}
For this task we evaluate attribute-based retrieval using the Claude-3-Sonnet model with both filtered and comprehensive settings. Additionally, we examine embedding-based retrieval using all other models. For embedding-based retrieval, we set $k=10$, meaning that 10 memory instances are retrieved (as opposed to 144 in the baseline).

\paragraph{Experimental Results}
Table~\ref{cr_res} shows the results for conversational recommendation evaluating comprehensive setting, attribute-based retrieval and embedding-based retrieval. As shown in the table, comprehensive memory augmentation tends to outperform the baseline and LLM-REDIAL model for recall and NDCG metrics. For genre match we find the results to be comparable when considering all attributes. However, attributed-based filtering retrieval still outperforms the LLM-REDIAL model and is comparable to the baseline with almost 90\% less memory retrieved.

Table \ref{cr_llm_res} presents the results of subjective LLM-based evaluation for Persuasiveness and Relatedness. The findings indicate that memory augmentation enhances partial persuasiveness by 10–11\% using both comprehensive and attribute-based retrieval, while also reducing unpersuasive recommendations and increasing highly persuasive ones by 4\% in attribute-based retrieval. Furthermore, the results highlights the effectiveness of embedding-based retrieval, which leads to a 12\% increase in highly persuasive recommendations and enhances all relatedness metrics. This illustrates how \mymethod enriches the recommendation process by incorporating condensed, relevant knowledge, thereby producing more persuasive and related recommendations. However, these improvements were not reflected in recall and NDCG metrics. 

\begin{table*}[]
\scriptsize
\centering
\begin{tabular}{lccccccc}
\toprule
\multicolumn{1}{c}{\small \textbf{Model}}     & \begin{tabular}[c]{@{}c@{}} \textbf{Avg. Items} \\   \textbf{Retrieved}\end{tabular} & \multicolumn{3}{c}{ \textbf{LLM-Persuasiveness} \%}                                  & \multicolumn{3}{c}{ \textbf{LLM-Relatedness}\%}                           \\ \midrule
 & \multicolumn{1}{l}{}                                                                  &  Unpers* & \begin{tabular}[c]{@{}c@{}}  Partially  Pers.\end{tabular} & \begin{tabular}[c]{@{}c@{}}  Highly  Pers.\end{tabular} &  Not Comp* & \multicolumn{1}{c}{ Comp} &  Match \\ \midrule
Baseline (Claude-3-Sonnet)                                  & 144                                                                                   & 16.0                   & 64.0                            & 13.0                         & 57.0                & 41.0                               & 2.0              \\
\hline
\multicolumn{8}{l}{\textbf{Attribute-Based Retrieval}} \\

\mymethod (Claude-3-Sonnet) & 15   & 2.0    & \textbf{75.0}  & 17.0     & 40.5   & 54.0 & 2.0   \\

 \hline
\multicolumn{8}{l}{\textbf{Embedding-Based Retrieval}} \\
\mymethod (Llama v3) & 10                                                              & 11.3  & 63.0                                                      & 20.4                                               & 19.3 & 80.1         & 0.5 \\
\mymethod (Mistral v1) & 10                                                              & 16.3 & 61.2   & 18.0    & \textbf{16.3} & \textbf{82.5} & \textbf{5.0 }   \\

\mymethod (Claude-3-Haiku)  & 10   & \textbf{1.6}   & 53.0 & \textbf{25.0 } & 23.3         & 74.4         & 2.2          \\ 

\mymethod (Claude-3-Sonnet)                             & 10                                                                                    & 2.0                       & 59.5                          & 20.0                 & 29.5      & 68.0   & 2.5   \\

\hline
\multicolumn{8}{l}{\textbf{Comprehensive}} \\
\mymethod (Claude-3-Sonnet)                                         & 144                                                                                   & 2.0                       & 74.0                             & 12.0                          & 42.5                 & 56.0                                 & 1.0           \\
\bottomrule
\end{tabular}
\small \caption{ \label{cr_llm_res}\small
Movie Recommendations results (with similar settings to Table \ref{cr_res}) using LLM-based metrics; (1) Persuasiveness— \% of Unpersuasive (lower is better), Partially, and Highly Persuasive cases. (2) Relatedness— \% of Not Comparable (lower is better), Comparable, and Exactly Matching cases. Best results are in bold. Comprehensive setting includes \textbf{ALL} augmentations. Totals may NOT sum to 100\% due to cases the LLM model could not evaluate.}
\end{table*}

\vspace{-2mm}
\subsection{Event Summarization}
\begin{figure}
    \centering
    \includegraphics[width=\linewidth]{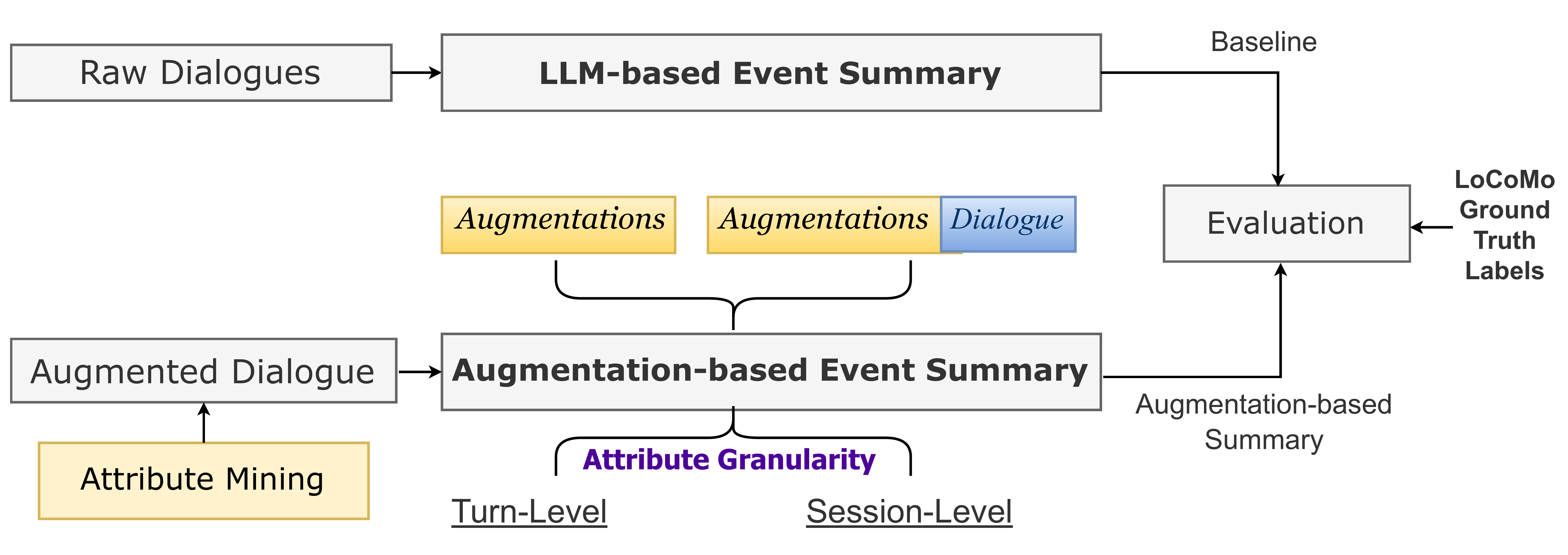}
    \small \caption{\small Evaluation framework for event summarization with \mymethod, exploring augmentation at Turn and Session levels, considering attributes alone or both attributes and dialogues for richer summaries. 
    }
    \label{fig:event_eval}
    
\end{figure}

We evaluate the effectiveness of \mymethod in enriching raw dialogues with relevant insights for event summarization. We utilize the generated annotations to identify key events within conversations and hence use them for event summarization. We compare the generated summaries against LoCoMo's event labels as the baseline. Figure~\ref{fig:event_eval} illustrates the experimental framework, where the baseline is the raw dialogues sent to the LLM model to generate an event summary, then both event summaries, from raw dialogues and augmentation based summaries, are compared to the ground truth summaries in the LoCoMo dataset.

\paragraph{Memory Augmentation}
In this experiment, we evaluate the effectiveness of augmentation granularity; turn-level dialogue augmentations as opposed to session-level dialogue annotations. We additionally, consider studying the effectiveness of using only the augmentations to generate the event summaries as opposed to using both the augmentations and their corresponding dialogue content.

\begin{table*}[]
\scriptsize
\centering
\begin{tabular}{llll|ccc|ccc|ccc}
\hline
\textbf{ Model}          
& \multicolumn{3}{c|}{\textbf{ Claude-3-Sonnet}}   & \multicolumn{3}{c|}{\textbf{ Llama v3}}           & \multicolumn{3}{c|}{\textbf{  Mistral v1}}         & \multicolumn{3}{c}{\textbf{  Claude-3-Haiku}}              \\ \hline
\textbf{}                & Rel.          & Coh.          & Con.          & Rel.          & Coh.          & Con.          & Rel.          & Coh.          & Con.          & Rel.          & Coh.          & Con.                   \\ \hline
Baseline Summary         & 3.27          & \textbf{3.52} & 2.86          & 2.03          & 2.64          & 2.68          & 3.39          & 3.71          & 4.10           & 4.00             & 4.4           & 3.83                   \\
 \mymethod(TL)    & 3.08          & 3.33          & 2.76          & 1.57          & 2.17          & 1.95          & 2.54          & 2.53          & 2.49          & 3.93          & 4.3           & 3.59                   \\
 \mymethod (SL)    & 3.08          & 3.39          & 2.68          & 2.0           & 2.62          & 3.67          & 4.13          & 4.41          & 4.29          & 3.96          & 4.30           & 3.77                   \\ \hline
 \mymethod+Dialogues (TL) & \textbf{3.29} & 3.46          & \textbf{2.92} & \textbf{2.45} & 2.19          & 2.87          & \textbf{4.30} & \textbf{4.53} & \textbf{4.60} & \textbf{4.23} & \textbf{4.52} & \textbf{4.16}\\
\mymethod+Dialogues (SL) & 3.05          & 3.41          & 2.69          & 2.24          & \textbf{2.80} & \textbf{3.86} & 4.04          & 4.48         & 4.33          & 3.93          & 4.33          & 3.73                   \\ \hline
\end{tabular}
\small \caption{\label{es_geval}
\small Event Summarization results using G-Eval metrics (higher is better): Relevance, Coherence, and Consistency. Comparing summaries generated with augmentations only at Turn-Level (TL) and Session-Level (SL) and summaries generated using both augmentations and dialogues (\mymethod+Dialogues) at TL and SL. Best results are in bold.
}

\end{table*}

\paragraph{Experimental Results}
As shown in Table \ref{es_geval}, our \mymethod model achieves performance comparable to the baseline, despite relying only on dialogue turns or sessions containing the event label. Notably, turn-level augmentations provided more precise and detailed event information, leading to improved performance over both the baseline and session-level annotations. 

For Claude-3-Sonnet, all metrics remain comparable, indicating that memory augmentations effectively capture the semantics within dialogues at both the turn and session levels. This proves that the augmentations sufficiently enhance context representation for generating event summaries. 
To further investigate how backbone LLMs impact augmentation quality, we employed Claude-3-Sonnet as opposed to Llama v3 for augmentation while still using Llama for event summarization. As presented in Table \ref{cl-ll}, Sonnet augmentations resulted in improved performance for all metrics, providing empirical evidence for the effectiveness and stability of Sonnet in augmentation. Additional experiments and detailed analysis are provided in Appendix~\ref{add_exp}.

\begin{table}[]
\small
\centering
\begin{tabular}{llccc}
\hline
\multicolumn{2}{l}{\textbf{Model}}             & \multicolumn{3}{c}{\textbf{G-Eval \% (↑)}}    \\ \hline
\multicolumn{2}{c}{}                           & Rel.          & Coh.          & Con.          \\ \hline
\multicolumn{2}{l}{Baseline(Llama v3 )}        & 2.03          & 2.64          & 2.68          \\
\multicolumn{2}{l}{Llama v3 + Llama v3}        & 2.45          & 2.19          & 2.87          \\
\multicolumn{2}{c}{Claude-3-Sonnet + Llama v3} & \textbf{3.15} & \textbf{3.59} & \textbf{3.17} \\ \hline
\end{tabular}
\small \caption{\label{cl-ll} \small Results for Event Summarization using Llama v3, where the baseline is the model without augmentation as opposed to the augmentation model (turn-level) using Claude-3-Sonnet vs Llama v3. \vspace{-3mm}}
\end{table}

\paragraph{Qualitative Analysis}
To more rigorously assess the quality of the autonomously generated augmentations, we conduct a qualitative analysis of the annotations produced by Claude-3 Sonnet. Using the DeepEval hallucination metric~\cite{deepeval}, we find that 99.14\% of the annotations are grounded in the dialogue, demonstrating a high level of factual consistency. The remaining 0.86\% primarily consist of abstract or generic attributes, rather than explicit inaccuracies. Additional experimental details and examples are provided in Appendix~\ref{qa}.

\section{Conclusion}

This paper introduced \mymethod, an autonomous memory augmentation framework that enhances LLM agents’ memory through structured, attribute-based augmentations. While maintaining competitive performance on standard metrics, \mymethod achieves substantial improvements in LLM-based evaluation scores, demonstrating its effectiveness in capturing semantic relevance and improving performance across tasks and datasets.
Experimental results show that both attribute-based filtering and embedding-based retrieval methods effectively leverage the generated augmentations. Priority-based augmentation, in particular, improves similarity search and retrieval accuracy. \mymethod also complements traditional RAG models by enabling customized, attribute-guided retrieval, enhancing the integration of memory with LLM reasoning.
Moreover, in benchmark comparisons, \mymethod consistently outperforms baseline models in overall accuracy and delivers stronger performance in recommendation tasks, yielding more persuasive outputs.
Qualitative analysis further confirms the high factual consistency of the generated annotations.
These results highlight \mymethod's potential as a scalable memory solution for LLM agents.

\section{Limitations}

While \mymethod demonstrates strong performance across multiple tasks and datasets, several limitations remain and highlight areas for future exploration. Although the model autonomously generates augmentations, it may occasionally produce abstract or overly generic annotations, especially in ambiguous dialogue contexts. While these are not factually incorrect, they may reduce retrieval specificity in tasks requiring fine-grained memory access.
Additionally, \mymethod’s performance is dependent on the capabilities of the underlying LLM used for attribute generation. Less capable or unaligned models may produce less consistent augmentations.
We also acknowledge that our current implementation is limited to text-based interactions. Future work could extend \mymethod to support multimodal inputs, such as images or audio, enabling richer and more comprehensive contextual representations.

\bibliography{custom}

\break
\appendix
\label{sec:appendix}

\section{Ethical Consideration}
We have thoroughly reviewed the licenses of all scientific artifacts, including datasets and models, ensuring they permit usage for research and publication purposes. To protect anonymity, all datasets used are de-identified. Our proposed method demonstrates considerable potential in significantly reducing both the financial and environmental costs typically associated with enhancing large language models. By lessening the need for extensive data collection and human labeling, our approach not only streamlines the process but also provides an effective safeguard for user and data privacy, reducing the risk of information leakage during training corpus construction. Additionally, throughout the paper-writing process, Generative AI was exclusively utilized for language checking, paraphrasing, and refinement.

\begin{figure*}[h]
    \centering
    \includegraphics[width=0.7\linewidth]{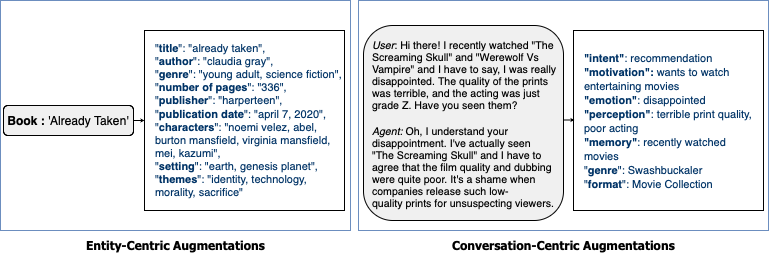}
    \caption{
    An example of entity-centric augmentation for the book 'Already Taken', and a  conversation-centric augmentation for a sample dialogue from the LLM-REDIAL dataset.
     }
    \label{fig:aug_ex}
\end{figure*}

\begin{figure*}
    \centering
    \includegraphics[width=0.7\linewidth]{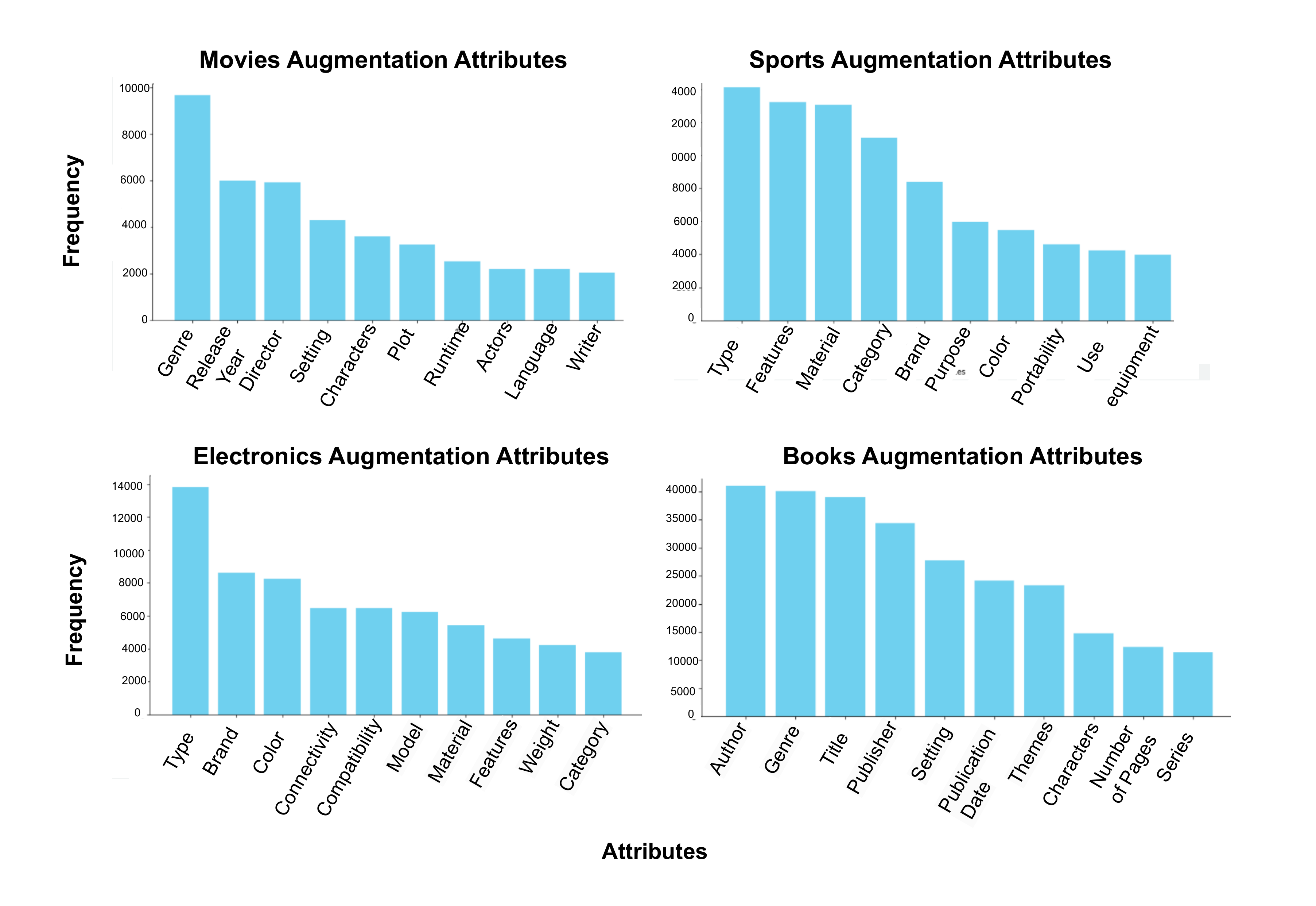}
    \caption{Top 10 attributes by frequency in the LLM-REDIAL dataset across domains (Movies, Sports Items, Electronics, and Books) using \mymethod Attribute Mining. Frequency indicates how often each attribute was generated to augment different movies.}
    \label{fig:all_att}
\end{figure*}

\begin{figure*}[]
    \centering
    \includegraphics[width=0.6\linewidth]{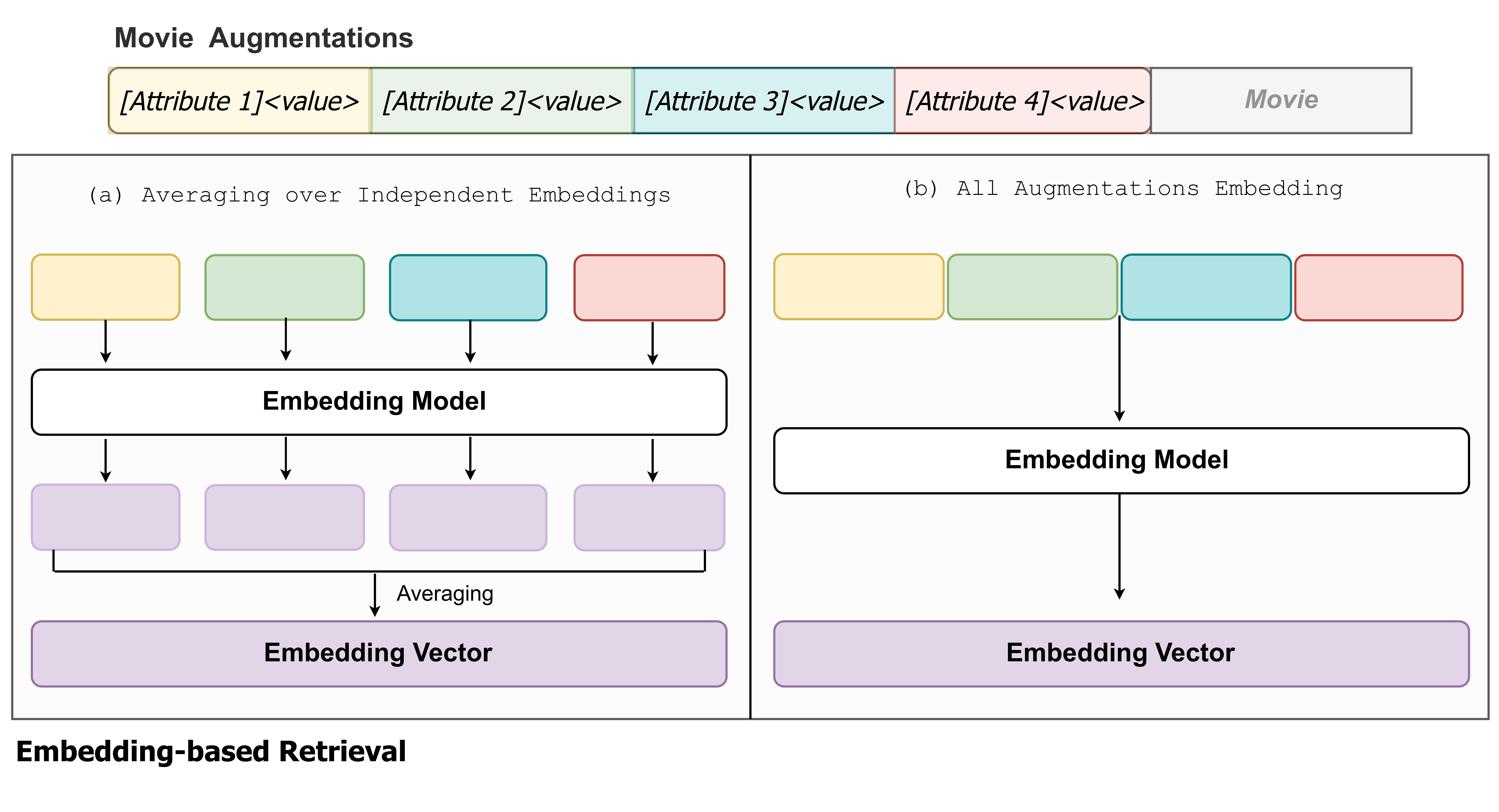}
    \caption{Embedding methods for Embedding-based retrieval methods using generated Movie augmentations including (a) Averaging over Independent Embeddings and (b) All Augmentations Embedding. }
    \label{fig:emb_ret}    
\end{figure*}

\section{Autonomous Memory Augmentation}
\subsection{Attribute Mining}
Figure \ref{fig:aug_ex} illustrates examples for the two types of attribute augmentation: entity-centric and conversation-centric. The entity-centric augmentation represents the main attributes generated for the book entitled 'Already Taken', where attributes are derived based on entity-specific characteristics such as genre, author, and thematic elements. The conversation-centric example illustrates the augmentation generated for a sample two turns dialogue from the LLM-REDIAL dataset, highlighting attributes that capture contextual elements such as user intent, motivation, emotion, perception, and genre of interest.

Furthermore, Figure \ref{fig:all_att} presents an overview of the top five attributes across different domains in the LLM-REDIAL dataset. These attributes represent the predominant attributes specific to each domain, highlighting the significance of different attributes in augmentation generation. Consequently, the integration of priority-based embeddings has led to improved performance.

\section{Embedding-based Retrieval}
\label{emb_ret}

\begin{figure*}[h]
    \centering
    \includegraphics[width=1\linewidth]{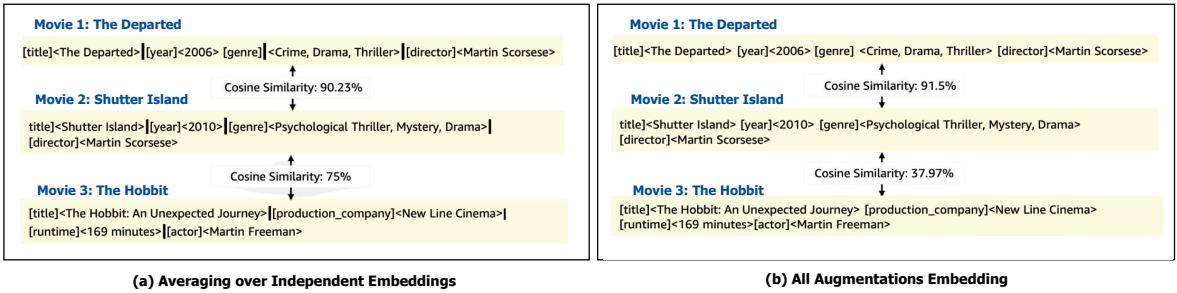}
    \caption{An illustrative example of augmentation embedding methods for three movies: (1) The Departed, (2) Shutter Island, and (3) The Hobbit. Movies 1 and 2 share similar attributes, whereas movies 1 and 3 differ. Te top 5 attributes of every movie were selected for a  simplified illustration. }
    \label{fig:cos_sim}
\end{figure*}

In the context of embedding-based memory retrieval, movies are augmented using \mymethod, and the generated attributes are embedded to retrieve relevant movies from memory. Two main embedding methods were considered:
\paragraph{(1) Averaging Over Independent Embeddings}
Each attribute and its corresponding value in the generated augmentations is embedded independently. The resulting attribute embeddings are then averaged across all attributes to generate the final embedding vector representation, as illustrated in Figure \ref{fig:emb_ret} which are subsequently used in similarity search to retrieve relevant movies.

\paragraph{(2) All Augmentations Embedding}
In this method, all generated augmentations, including all attributes and their corresponding values, are encoded into a single embedding vector and stored for retrieval as shown in Figure \ref{fig:emb_ret}. Additionally, Figure \ref{fig:cos_sim} presents the cosine similarity results for both methods. As depicted in the figure, averaging over all augmentations produces a more consistent and reliable measure, as it comprehensively captures all attributes and effectively differentiates between similar and distinct characteristics. Consequently, this method was adopted in our experiments.

\begin{table*}[htb]
\centering
\footnotesize{}
\begin{tabular}{p{15cm}}
\hline
\textbf{Question Augmentation}       \\ \hline
Given the following question, determine what are the main inquiry attribute to look for and the person the question is for. Respond in the format: Person:{[}names{]}Attributes:{[}{]}.
\\ \hline
\textbf{Basic Augmentation}           \\ \hline
You are an expert annotator who generates the most relevant attributes in a conversation. Given the conversation below, identify the key attributes and their values on a turn by turn level.\\ Attributes should be specific with most relevant values only. Don't include speaker name. Include value information that you find relevant and their names if mentioned. Each dialogue turn contains a dialogue id between {[} {]}. Make sure to include the dialogue the attributes and values are extracted form. Important: Respond only in the format {[}\{speaker name:{[}Dialog id{]}:{[}attribute{]}\textless{}value\textgreater{}\}{]}.\\ Dialogue Turn:\{\}
\\ \hline
\textbf{Priority Augmentation}
\\ \hline
You are an expert dialogue annotator, given the following dialogue turn generate a list of attributes and values for relevant information in the text.\\ Generate the annotations in the format: {[}attribute{]}\textless{}value\textgreater where attribute is the attribute name and value is its corresponding value from the text.\\ and values for relevant information in this dialogue turn with respect to each person. Be concise and direct.\\ Include person name as an attribute and value pair.\\ Please make sure you read and understand these instructions carefully.\\ 1- Identify the key attributes in the dialogue turn and their corresponding values.\\ 2- Arrange attributes descendingly with respect to relevance from left to right.\\ 3- Generate the sorted annotations list in the format: {[}attribute{]}\textless{}value\textgreater where attribute is the attribute name and value is its corresponding value from the text.\\ 4- Skip all attributes with none vales\\ Important: YOU MUST put attribute name is between {[} {]} and value between \textless \textgreater{}. Only return a list of {[}attribute{]}\textless{}value\textgreater nothing else. Dialogue Turn: \{\} \\ \hline
\end{tabular}
\caption{\label{qa_prmpt} Prompts used in Question Answering for generating augmentations for questions. Also, augmentations for conversations, utilizing both basic and priority augmentations. }
\end{table*}

\section{Question Answering}

\subsection{Prompts}
Table \ref{qa_prmpt} outlines the prompts used in the Question Answering task for generating augmentations in both questions and conversations.

\begin{table*}[htb]
\centering
\footnotesize{}
\begin{tabular}{p{15cm}}
\hline
\textbf{Basic Augmentation}       \\ \hline
For the following movie identify the most important attributes independently. Determine all attributes that describe the movie 
based on your knowledge of this movie. Choose attribute names that are common characteristics of movies in general.
Respond in the following format:
[attribute]<value of attribute>.
The Movie is: \{\}
\\
\\ \hline
\textbf{Priority Augmentation}           \\ \hline
You are a movie annotation expert tasked with analyzing movies and generating key-attribute pairs.
For the following movie identify the most important. Determine all attribute that describe the movie based on your knowledge of this movie.
Choose attribute names that are common characteristics of movies in general.
Respond in the following format:
[attribute]<value of attribute>.
\textit{Sort attributes from left to right based on their relevance.}
The Movie is:\{\}
\\ \hline
\textbf{Dialogue Augmentation}
\\ \hline
Identify the key attributes that best describe the movie the user wants for recommendation in the dialogue.
These attributes should encompass movie features that are relevant to the user sorted descendingly with respect to user interest.
Respond in the format: [attribute]<value>.
\\ \hline
\end{tabular}
\caption{\label{cr_prmpt} Prompts used in Conversational Recommendation for recommending Movies utilizing both basic and priority augmentations. }
\end{table*}

\section{Conversational Recommendation}
\subsection{Prompts}
Table \ref{cr_prmpt} presents the prompts used in Conversational Recommendation for movie recommendations, incorporating both basic and priority augmentations. 

\subsection{Evaluation Framework}
\begin{figure*}[]
    \centering
    \includegraphics[width=0.8\linewidth]{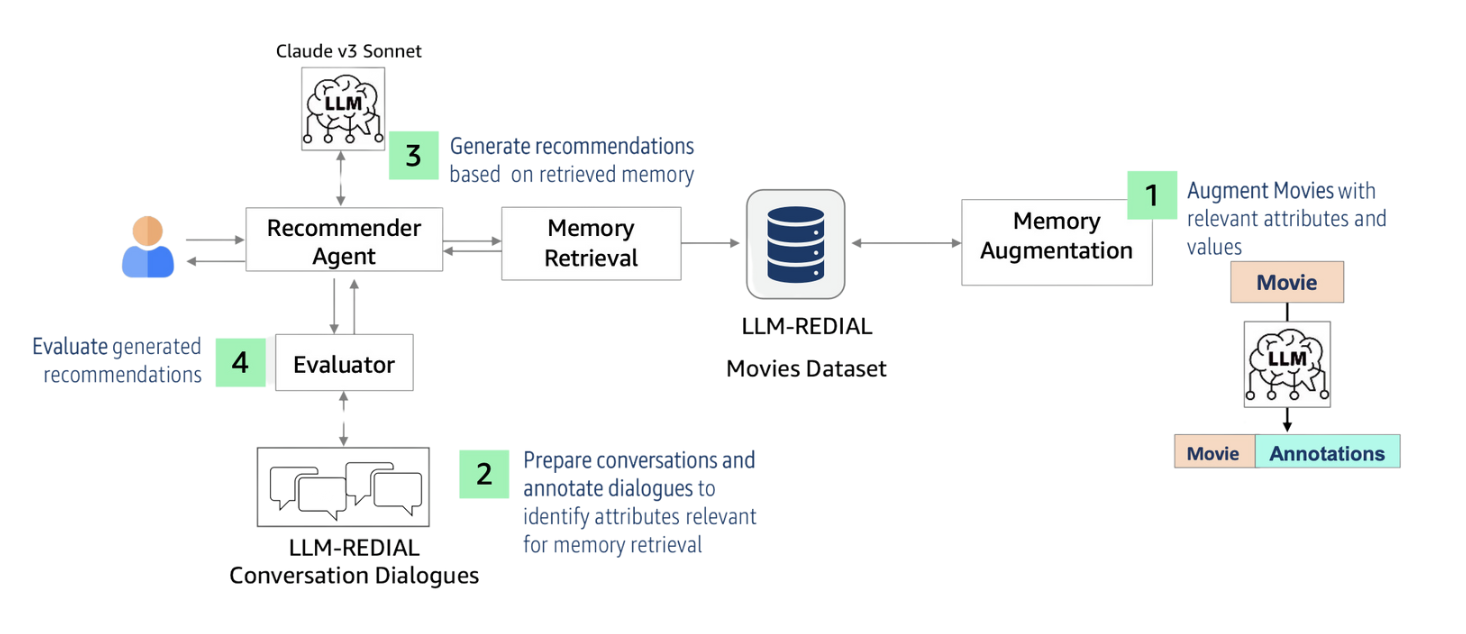}
    \caption{Evaluation Framework for Conversation Recommendation Task.}
    \label{fig:eval_cr}
\end{figure*}

Figure \ref{fig:eval_cr} presents the evaluation framework for the Conversation Recommendation task. The process begins with (1) augmenting all movies in memory using entity-centric augmentations to enhance retrieval effectiveness. (2) Next, all dialogues in the dataset are prepared to simulate the recommendation process by masking the ground truth labels and prompting the LLM to find the masked labels based on augmentations from previous user interactions. (3) Recommendations are then generated using the retrieved memory, which may be attribute-based—for instance, filtering movies by specific attributes such as genre or using embedding-based retrieval. (4) Finally, the recommended movies are evaluated against the ground truth labels to assess the accuracy and effectiveness of the retrieval and recommendation approach.

\begin{table*}[htb]
\centering
\footnotesize{}
\begin{tabular}{p{15cm}}
\hline
\textbf{Dialogue Augmentation}       \\ \hline
Given the following attributes and values that annotate a dialogue for every speaker in the format [attribute]<value>, generate a summary for the event attributes only to describe the main and important events represented in these annotations. Refrain from mentioning any minimal event. Include any event-related details and speaker. Format: a bullet paragraph for major life events for every speaker with no special characters. Don't include anything else in your response or extra text or lines. Don't include bullets. Input annotations: \{\}
\\ \hline
\end{tabular}
\caption{\label{es_prmpt} Prompt used in Event Summarization to augment dialogues }
\end{table*}

\begin{table*}
\scriptsize
\centering
\begin{tabular}{l|ccc|ccc|ccc|ccc}
\hline
\textbf{Model}              & \multicolumn{3}{c|}{\textbf{Llama v3}}                       & \multicolumn{3}{c|}{\textbf{Mistral v1}}                     & \multicolumn{3}{c|}{\textbf{Claude-3 Haiku}}                 & \multicolumn{3}{c}{\textbf{Claude-3 Sonnet}}                 \\ \hline
                            & \textbf{Rel.} & \textbf{Coh.} & \textbf{Con.} & \textbf{Rel.} & \textbf{Coh.} & \textbf{Con.} & \textbf{Rel.} & \textbf{Coh.} & \textbf{Con.} & \textbf{Rel.} & \textbf{Coh.} & \textbf{Con.} \\ \hline
Baseline LLM Summary        & 2.23 & 2.66 & 2.63 & 3.34 & 3.77 & 4.11 & 3.97 & 4.33 & 3.79 & \textbf{3.27} & \textbf{3.64} & 2.78 \\ \hline
MemInsight (TL)             & 1.60 & 2.17 & 1.95 & 2.53 & 2.49 & 2.38 & 3.98 & 4.37 & 3.66 & 3.09 & 3.27 & 2.77 \\ \hline
MemInsight (SL)             & 1.80 & 2.62 & 3.67 & 4.09 & 4.38 & 4.19 & 3.94 & 4.31 & 3.69 & 3.08 & 3.39 & 2.68 \\ \hline
MemInsight + Dialogues (TL) & \textbf{2.41} & \textbf{2.79} & 3.01 & \textbf{4.30} & \textbf{4.53} & \textbf{4.60} & \textbf{4.24} & \textbf{4.43} & \textbf{4.16} & 3.25 & 3.43 & \textbf{2.86} \\ \hline
MemInsight + Dialogues (SL) & 2.01 & 2.70 & \textbf{3.86} & 4.04 & 4.48 & 4.34 & 3.95 & 4.33 & 3.71 & 3.02 & 3.37 & 2.73 \\ \hline
\end{tabular}
\caption{\label{raw_summ} LLM-based evaluation scores for event summarization using relevance (Rel.), coherence (Coh.), and consistency (Con.) across different models and augmentation settings. Baseline summaries are generated using zero-shot prompting without memory augmentation. MemInsight is evaluated in both turn-level (TL) and session-level (SL) configurations, with and without access to dialogue context.}
\end{table*}

\subsection{Event Summarization}
\subsubsection{Prompts}
Table \ref{es_prmpt} presents the prompt used in Event Summarization to augment dialogues by generating relevant attributes. In this process, only attributes related to events are considered to effectively summarize key events from dialogues, ensuring a focused and structured summarization approach.

\begin{figure*}
    \centering
    \includegraphics[width=\linewidth]{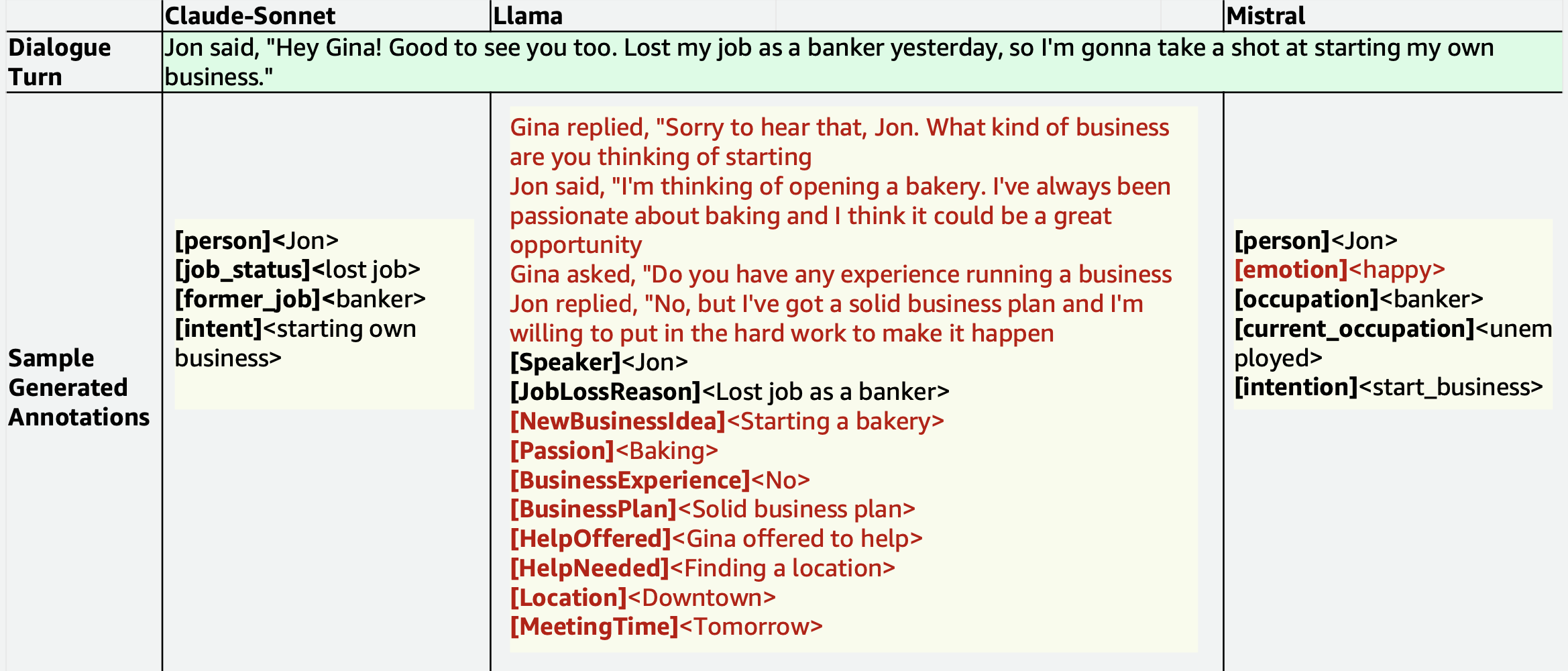}
    \caption{Augmentation generated on a Turn-level for a sample dialogue turn from the LoCoMo dataset using Claude-3-Sonnet, Llama v3 and Mistral v1 models. }
    \label{fig:1_ann}
\end{figure*}

\begin{figure*}
    \centering
    \includegraphics[width=\linewidth]{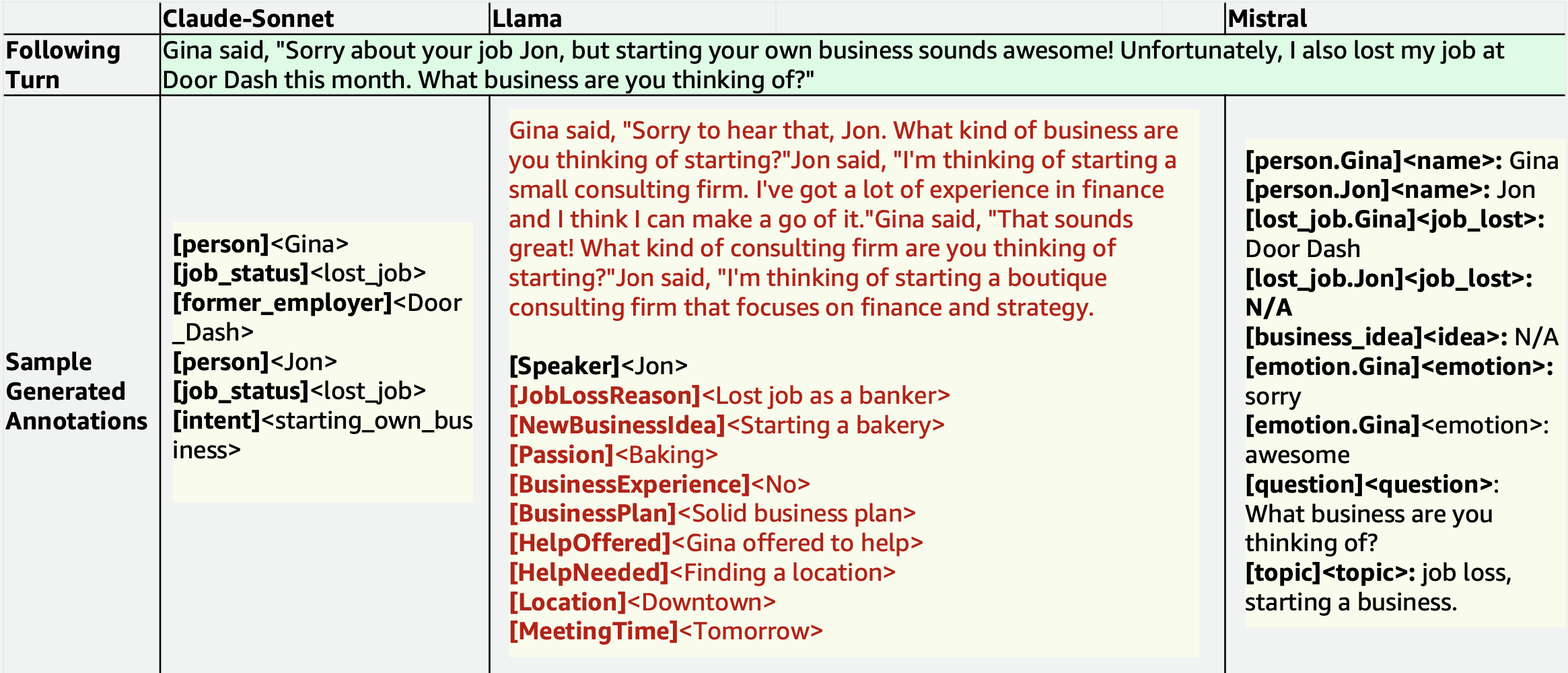}
    \caption{Augmentations generated for the turn following the turn in Figure \ref{fig:1_ann}} using Claude-3-Sonnet, Llama v3 and Mistral v1 models. Hallucinations are presented in red.
    \label{fig:2_ann}
\end{figure*}

\subsection{Additional Experiments}
\label{add_exp}
In this experiment, we include an additional baseline for event summarization: raw summaries generated directly by LLMs using zero-shot prompting, without any memory augmentation. This serves as a clear reference point to isolate the impact of \mymethod’s augmentation strategy on summarization quality. Table \ref{raw_summ} shows the results of this experiment. As illustrated, \mymethod consistently improves event summarization quality across models, with the best performance achieved when augmentations are integreted with dialogue context highlighting the value of fine-grained annotations and contextual grounding. Overall, the findings confirm that \mymethod enhances the factual and semantic quality of generated summaries.

\section{Qualitative Analysis}
\label{qa}
Figure \ref{fig:1_ann} illustrates the augmentations generated using different LLM models, including Claude-Sonnet, Llama, and Mistral for a dialogue turn from the LoCoMo dataset. As depicted in the figure, augmentations produced by Llama include hallucinations, generating information that does not exist. In contrast, Figure \ref{fig:2_ann} presents the augmentations for the subsequent dialogue turn using the same models. Notably, Claude-Sonnet maintains consistency across both turns, suggesting its stable performance throughout all experiments. While Mistral model tend to be less stable as it included attributes that are not in the dialogue.
A hallucination evaluation conducted using DeepEval yielded a score of 99.14\%, indicating strong factual consistency. Table~\ref{tab:low_hal} presents examples of annotations with lower scores. While these annotations are more generic or abstract, they remain semantically aligned with the original input.

 \begin{table}[t]
\renewcommand{\arraystretch}{1.2}
\centering
\footnotesize
\begin{tabular}{|p{7cm}|p{6cm}|p{2cm}|}
\hline
\textbf{Input} & \textbf{Augmentations} & \textbf{Hall. Score} \\
\hline
\texttt{'Evan': [[``Evan's son had an accident where he fell off his bike last Tuesday but is doing better now.", D20:3], [``Evan is supportive and encouraging towards Sam, giving advice to believe in himself and take things one day at a time.", D20:9], [``Evan is a painter who finished a contemporary figurative painting emphasizing emotion and introspection.", D20:15], [``Evan had a painting published in an exhibition with the help of a close friend.", D20:17]], 'Sam': [[``Sam used to love hiking but hasn't had the chance to do it recently.", D20:6], [``Sam is struggling with weight and confidence issues, feeling like they lack motivation.", D20:8], [``Sam acknowledges that trying new things can be difficult.", D20:12]]} & 

\texttt{"evan":\{"[event]":"<son's accident>",
"[emotion]":"<worry>",
"[hobby]":"<hiking>",
"[activity]":"<painting>"\}, "sam":\{"[emotion]":"<struggling>", "[issue]":"<weight>", "[emotion]":"<lack of confidence>", "[action]":"<trying new things>"\}\}} & 
0.66 \\
\hline
\texttt{\{'James': [[``James has a dog named Ned that he adopted and can't imagine life without.", D21:3], [``James is interested in creating a strategy game similar to Civilization.", D21:9], [``James suggested meeting at Starbucks for coffee with John.", D21:13]], 'John': [[``John helps his younger siblings with programming and is proud of their progress", D21:2], [``John is working on a coding project with his siblings involving a text-based adventure game.", D21:6], [``John prefers light beers over dark beers when going out.", D21:16], [``John agreed to meet James at McGee's Pub after discussing different options.", D21:18]]\}} & 
\texttt{\{"james":\{"[emotion]":"<excited>", "[intent]":"<socializing>", "[topic]":"<dogs>", "[topic]":"<gaming>", "[topic]":"<starbucks>", "[topic]":"<pubMeeting>", "[activity]":"<coffee>", "[activity]":"<beer>"\}, "john":\{"[topic]":"<siblings>", "[topic]":"<programming>", "[activity]":"<adventure game>", "[emotion]":"<proud>", "[intent]":"<socializing>"\}\}} & 
0.50 \\
\hline
\end{tabular}
\caption{\mymethod annotations that scored below 1\% hallucination rate in the DeepEval hallucination evaluation.}
\label{tab:low_hal}
\end{table}

\end{document}